\documentclass[letterpaper, 10 pt, conference]{ieeeconf}  
\IEEEoverridecommandlockouts   

\usepackage{graphicx}  
\usepackage{pdflscape}
\usepackage{rotating}
\usepackage{booktabs}  
\usepackage{amssymb}   
\usepackage{array}     
\usepackage{xcolor}    
\usepackage{amsmath}   
\usepackage{makecell} 
\usepackage{epstopdf}
\usepackage{subcaption}
\usepackage{color}
\usepackage{multirow}
\usepackage{adjustbox}
\usepackage{caption}
\usepackage{url}
\usepackage{amsmath}
\usepackage{cite}
\usepackage{bbding}

\title{\LARGE \bf 
GTAD: Global Temporal Aggregation Denoising Learning for 3D Semantic Occupancy Prediction
}

\author{
 Tianhao Li , Yang Li , Mengtian Li , Yisheng Deng , Weifeng Ge$^{*}$
\thanks{Tianhao Li, Yisheng Deng and Weifeng Ge are with the Department of Computer and Technology, Fudan University, Shanghai, China. *Corresponding author. E-mails: wfge@fudan.edu.cn}%
\thanks{Yang Li is with the School of Computer Science and Technology, East China Normal University, Shanghai,China. E-mails: yli@cs.ecnu.edu.cn}%
\thanks{Mengtian Li is with the Shanghai Film Academy of Shanghai University, Shanghai, China. E-mails: mtli@{shu,fudan}.edu.cn}%
}

\begin{document}
\maketitle
\thispagestyle{empty}
\pagestyle{empty}

\begin{abstract}

Accurately perceiving dynamic environments is a fundamental task for autonomous driving and robotic systems. Existing methods inadequately utilize temporal information, relying mainly on local temporal interactions between adjacent frames and failing to leverage global sequence information effectively.
To address this limitation, we investigate how to effectively aggregate global temporal features from temporal sequences, aiming to achieve occupancy representations that efficiently utilize global temporal information from historical observations. For this purpose, we propose a global temporal aggregation denoising network named GTAD, introducing a global temporal information aggregation framework as a new paradigm for holistic 3D scene understanding. Our method employs an in-model latent denoising network to aggregate local temporal features from the current moment and global temporal features from historical sequences. This approach enables the effective perception of both fine-grained temporal information from adjacent frames and global temporal patterns from historical observations. As a result, it provides a more coherent and comprehensive understanding of the environment. Extensive experiments on the nuScenes and Occ3D-nuScenes benchmark and ablation studies demonstrate the superiority of our method.

\end{abstract}


\section{INTRODUCTION}

In the field of autonomous driving, comprehensively understanding 3D environments is a fundamental and critical task. To build a safe and trustworthy autonomous driving system, it is essential to deeply explore and skillfully utilize detailed temporal sequence data. In recent years, camera-based Bird’s Eye View (BEV) technology has achieved significant breakthroughs in 3D environment understanding. By integrating multi-view visual information into a unified BEV framework, this technology has demonstrated its value across various applications, including BEV semantic segmentation\cite{zhou2022cross,milioto2020lidar,zhou2022cross}, 3D object detection\cite{hu2021fiery,dd3d,mmdet3d2020}and high-precision vectorized map construction\cite{liao2022maptr,roddick2020predicting}.

\begin{figure}[htbp]
    \centering
    \includegraphics[width=0.5\textwidth, keepaspectratio]{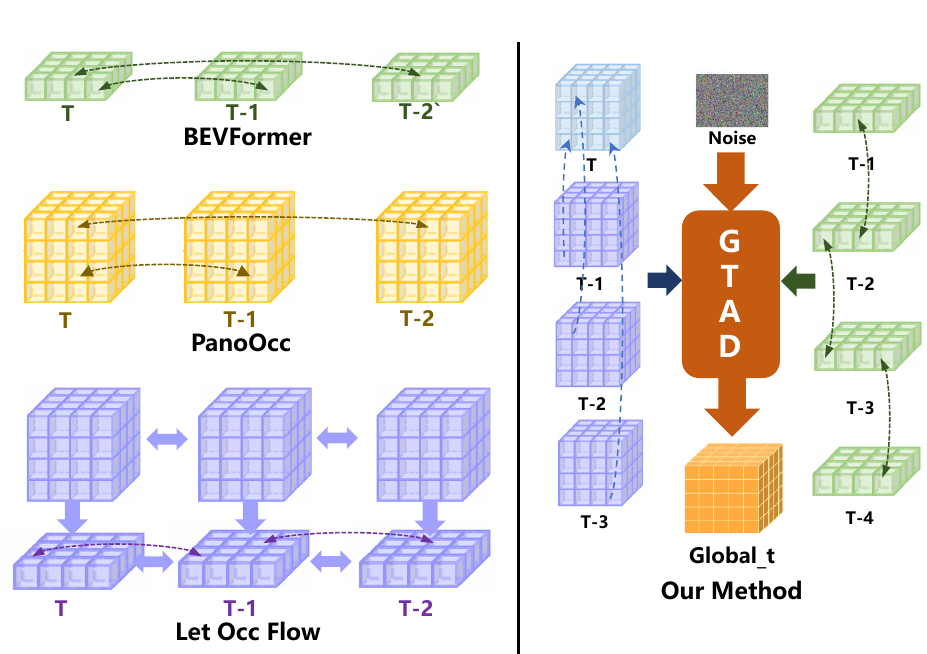} 
    \caption{Comparison of temporal aggregation methods. Our method combines local features at the current step and global features from the historical queue using denoising network for precise scene perception.}
    \label{fig:comare}
\end{figure}
Multi-camera 3D object detection plays a pivotal role in vision-driven 3D environmental perception\cite{liu2022spatial,li2021multi}. However, it faces challenges from long-tail distributions, which make it exceptionally difficult to recognize all object categories in real-world scenarios. To address this limitation, reconstructing the surrounding 3D scene has become crucial for enhancing downstream perception tasks.

Occupancy representation has garnered significant attention due to its ability to comprehensively and precisely depict diverse elements in scenes, including open-set objects, irregularly shaped entities, and unique road structures. Consequently, emerging methods \cite{miao2023occdepth,huang2023tri,wang2023openoccupancy,li2023voxformer} are increasingly focusing on dense semantic occupancy prediction, aiming to achieve finer-grained scene understanding and analysis.

In occupancy prediction tasks, temporal information plays a vital role in identifying occluded objects. However, most current multi-camera 3D detection methods fail to adequately exploit and integrate temporal sequence information, limiting their performance in dynamic scenario\cite{BEVFormer,Liu2024LetOF,wang2023panooccunifiedoccupancyrepresentation,xia2024henethybridencodingendtoend}. Earlier approaches attempted to address this by simply stacking BEV features across timestamps to leverage historical information\cite{huang2022bevdet4dexploittemporalcues,li2022bevstereoenhancingdepthestimation}. Given the rapid changes of objects in autonomous driving scenarios, however, such naive stacking introduces excessive computational costs and noise, which significantly degrades the detection accuracy and overall efficiency.

BEVformer\cite{BEVFormer} propagates temporal information cyclically from past to present via BEV features, employing temporal self-attention to aggregate historical BEV features and capture temporal dynamics. PanoOcc\cite{wang2023panooccunifiedoccupancyrepresentation} aligns temporal information by sampling voxel queries from historical timestamps and concatenating previous aligned voxel queries with current queries to obtain temporally aggregated queries \(\mathbf{Q}_t\).
However, the temporal aggregation in BEVformer and PanoOcc remains limited to local interactions between two timestamps, lacking the capability to globally integrate features across entire temporal sequences. Consequently, a more comprehensive spatiotemporal 3D scene understanding paradigm is required.

Inspired by diffusion models\cite{ho2020denoisingdiffusionprobabilisticmodels}, denoising learning has been widely applied in generative tasks. It can capture subtle details in the data and generate results with rich details, thereby effectively aggregating temporal information for finer-grained perception. 
Therefore, we adopt a new paradigm that formulates the 3D semantic occupancy prediction as a denoising learning process. Specifically, we introduce noise into BEV feature maps of the current time step to generate noisy BEV feature maps that incorporate global temporal information, thereby enhancing the model's ability to perceive global temporal information at a finer granularity.
However, directly applying diffusion models to the task of semantic occupancy prediction severely degrades prediction speed and consumes a significant amount of computational resources. 
Thus, we use the In-model latent denoising network\cite{zhou2025detrackinmodellatentdenoising},which can decompose the denoising learning process into individual denoising blocks within the model, rather than running the model multiple times. As a result, we summarize the 3D semantic occupancy prediction task as an in-model latent denoising learning process.

The denoising network takes two conditional inputs: 1) locally aggregated current voxel features and 2) globally aligned historical BEV features. It progressively removes noise from the coarsely perceived BEV map at the current timestamp, ultimately outputting a globally aware BEV map that integrates both local temporal details from adjacent frames and global historical patterns. This design enables real-time processing of occluded objects and historical information while maintaining computational efficiency.

In this work, we propose a global temporal aggregation framework called \textbf{GTAD}, which effectively enriches the model with global temporal context through denoising networks.Our key contributions are summarized as follows:
\begin{itemize}
\item We introduce an in-model latent denoising network for global temporal aggregation, enhancing 3D scene understanding. It aggregates local temporal information from adjacent frames and global patterns from historical sequences, enabling coherent and comprehensive perception.
\item We propose an attention-based global temporal information interaction mechanism that overcomes the isolated processing of historical features in prior methods, significantly enhancing the integration of global historical information.
\item On the Occ3D-nuScenes and nuScenes benchmark, our method achieves state-of-the-art performance for camera-based occupancy field prediction tasks
\end{itemize}

\begin{figure*}[htbp]
    \centering
    \includegraphics[width=\textwidth]{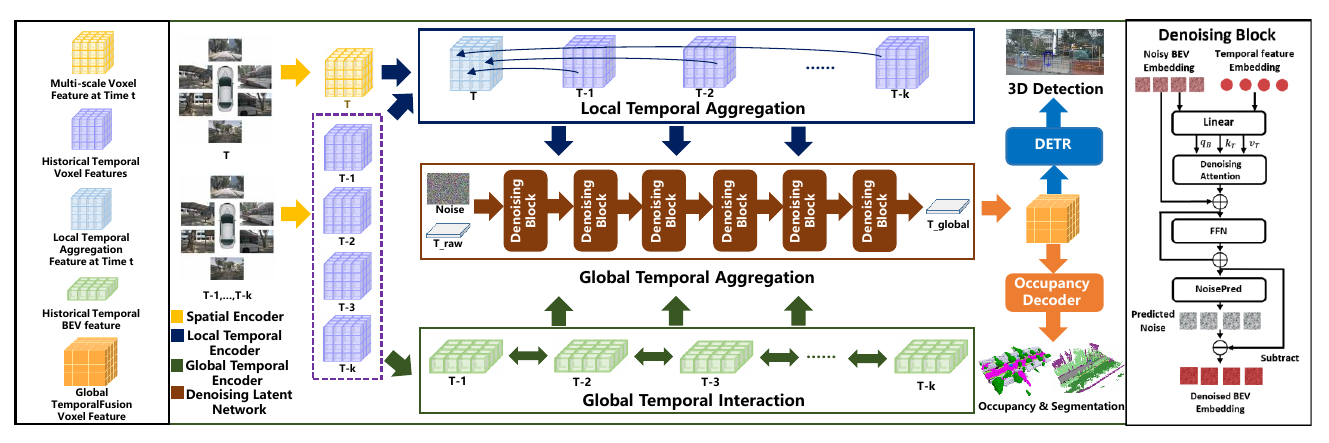}  
    \caption{GTAD framework starts with an image backbone network to extract multi-scale features from multi-view images across frames. Voxel queries then learn current voxel features via the Spatial Encoder. The Local Temporal Encoder align and combine previous voxel features with the current frame. The Global Temporal Encoder maps the voxel feature queue to BEV space and enables global temporal information interaction using time-decay weights. Local and global temporal aggregation features are embedded into the in-model latent denoising network to output the occupancy volume representation at time T. Finally, the 3D semantic occupancy results are generated by the Occupancy Decoder.}  
    \label{fig:pipeline}  
\end{figure*}

\section{ RELATED WORK}
\subsection{3D Semantic Occupancy Prediction}
In recent years, vision-based 3D occupancy prediction\cite{wang2023panooccunifiedoccupancyrepresentation,miao2023occdepth,li2023voxformer} has emerged as a promising method for perceiving environments. It aims to reconstruct 3D scenes from images and is divided into sparse prediction\cite{milioto2020lidar} and dense prediction\cite{tian2023occ3d}. TPVFormer\cite{huang2023tri} introduces a tri-perspective method for sparse prediction, MonoScene\cite{cao2022monoscene} uses a U-Net for dense prediction, and VoxFormer\cite{li2023voxformer} focuses on voxel query selection. Benchmarks like OpenOccupancy\cite{wang2023openoccupancy} and Occ3D\cite{tian2023occ3d} provide annotated datasets for evaluation and dynamic object modeling. These studies advance camera-based 3D scene understanding and benefit autonomous driving and robotic navigation.

\subsection{Temporal Aggregation}
In BEVFormer\cite{BEVFormer}, temporal fusion is achieved through the Temporal Self-Attention module, which integrates BEV features from different time steps by selecting three frames from the previous four. These features are concatenated with the current frame’s queries and fused via deformable attention\cite{zhu2020deformable} after alignment using sensor motion information. Four points are sampled per query pixel, and the current frame’s features are obtained by averaging the derived features, enhancing adaptability to dynamic scenes. In PanoOcc\cite{wang2023panooccunifiedoccupancyrepresentation}, the "Temporal Encoder" module processes temporal information by aligning and fusing voxel features from different time steps using the ego-vehicle’s pose for coordinate transformation and interpolation sampling. The aligned features are concatenated with the current features and fused via a residual 3D convolutional block, improving perception in dynamic scenes and model robustness.
HENet\cite{xia2024henethybridencodingendtoend} uses the AFFM module to aggregate BEV features from adjacent frames, improving the efficiency of temporal aggregation. Building on this, Let Occ Flow\cite{Liu2024LetOF}introduces the BFAM module to enhance the exchange of adjacent temporal information in BEV features and incorporates temporal information into voxel representations.

\subsection{In-model Latent Denoising Learning}
DDPM\cite{ho2020denoisingdiffusionprobabilisticmodels} enhance image quality and diversity via denoising diffusion learning but suffer from slow inference due to multiple decoder passes. DeTrack\cite{zhou2025detrackinmodellatentdenoising}introduces an in-model latent variable denoising learning paradigm for visual object tracking using a Denoising Vision Transformer (Denoising ViT) with multiple blocks to remove noise from bounding boxes in a single forward pass. This enables efficient real-time tracking, enhances stability and robustness via visual and trajectory memory, improves adaptability to unseen data, and reduces computational costs. This work show that denoising networks have broad applications in vision tasks, improving model robustness and generalization.

Due to its excellent performance in the field of feature aggregation and surprisingly low computational cost, in-model latent denoising network is applied as a method for global temporal feature aggregation, in contrast to traditional attention mechanisms that are often expensive and inefficient. This approach can effectively aggregate temporal information from the global time sequence under lower computational resource consumption, thereby significantly improving the performance of 3D semantic occupancy prediction.

\section{Global Temporal Aggregation Denoising Network}

\subsection{Overall Architecture and Problem Formulation}
We introduce GTAD, an in-model latent denoising network for 3D semantic occupancy. It takes multi-frame, multi-view images as input and outputs semantic occupancy maps. The method employs an image backbone to extract multi-scale features, which are subsequently processed by a condition encoder comprising spatial, local temporal, and global temporal encoders.
The spatial encoder extracts voxel features of the current moment without temporal information,
local temporal encoder preliminarily aggregates past temporal information into the features of the current moment,
and global temporal encoder facilitates global information interaction among historical features in the temporal queue, enabling each historical feature to preliminarily integrate past temporal information.
Finally, the current feature and the historical feature queue, which have preliminarily integrated past temporal information, as conditional inputs are embedded into the subsequent denoising network to produce the occupancy volume at time t.

Given multi-view images as input, the goal of the camera-based 3D semantic occupancy task is to predict a dense panoramic voxel volume surrounding the vehicle. Specifically, we use the multi-view images at the current time step, \( I_t = \{I_{1}^{t}, I_{2}^{t}, \dots, I_{n}^{t}\} \), as well as images from the previous \( k \) frames, \( I_{t-1}, \dots, I_{t-k} \), are used as inputs. Here, \( n \) denotes the index of the camera views, and \( k \) represents the number of historical frames.
The model outputs the semantic voxel volume \( Y_t \in \{w_0, w_1, \dots, w_C\}^{H \times W \times Z} \) for the current frame. 
Here, \( H \), \( W \), and \( Z \) denote the height, width, and depth of the surrounding voxel volume, respectively, \( C \) represents the total number of semantic classes in the scene, and \( w_0 \) indicates the empty voxel grid.

\subsection{Spatial Encoder}
Given voxel queries $\mathbf{Q}$ and extracted image feats $\mathbf{F}$ as input, the occupancy encoder outputs the fused voxel features $\mathbf{Q}_f \in \mathbb{R}^{H\times W\times Z \times D}$. $H$,$W$ and $Z$ represent the shape of the output voxel features, and $D$ is the embedding dimension.
The core difference lies in the choice of \emph{reference points} to generalize the BEV queries to voxel queries\cite{BEVFormer}.

Referencing the contributions of PanoOcc\cite{wang2023panooccunifiedoccupancyrepresentation}, for a voxel query $\mathbf{q}$ located at $(i, j, k)$, the process of Spatial cross-attention (SCA) can be formulated as follows:
\begin{equation}
    \text{SCA}(\mathbf{q},\mathbf{F}) = \frac{1}{|v|}\sum_{n\in v}\sum_{m=1}^{M_1} \text{DA} (\mathbf{q}, \mathbf{\pi_n}(\mathbf{Ref}_{i,j,k}^{m}),\mathbf{F}_n)
\end{equation}
\begin{equation}
    \mathbf{Ref}_{i,j,k}^{m} = (x_i^m,y_j^m,z_k^m)
\end{equation}

where \( n \) index the camera views, \( m \) index the reference points, and \( M_1 \) is the total number of sampling points for each voxel query. \( v \) is the set of image views for which the projected 2D point of the voxel query can fall on. \(\mathbf{F}_{n}\) denotes the image features of the \( n \)-th camera view. \(\mathbf{\pi_n}(\mathbf{Ref}_{i,j,k}^{m})\) represents the \( m \)-th projected reference point in the \( n \)-th camera view, projected by the projection matrix \(\mathbf{\pi_n}\) from the voxel grid located at \((i, j, k)\). \(\text{DA}\) stands for deformable attention. The real position of a reference point located at voxel grid \((i, j, k)\) in the ego-vehicle frame is \((x_i^m, y_j^m, z_k^m)\).


\subsection{Temporal Encoder}
In the local temporal encoder, we employ temporal self-attention to perform local temporal aggregation of the current moment's features, while in the global temporal encoder, we leverage a global temporal information communcication mechanism to allow each historical feature in the queue to preliminarily integrate past temporal information.

\noindent\textbf{Local Temporal Aggregation.}
Due to voxel features can capture fine details like shape and height for 3D structures, We use voxel features with local temporal information for local perception in the GTAD. Following BEVFormer~\cite{BEVFormer} and PanoOcc~\cite{wang2023panooccunifiedoccupancyrepresentation}, we use a Temporal Self-Attention (TSA) layer to integrate historical voxel features from adjacent timestamps into the current environment representation. Specifically, given voxel queries \( Q_t \) at timestamp \( t \), we locally align past queries \( \{Q_{t-1}, Q_{t-2}, \dots, Q_{t-k}\} \) with \( Q_t \) using the method of voxel alignment in 3D space. This alignment ensures that features on the same voxel grid correspond to the same real-world location. We denote the aligned historical voxel features as \( Q_{k} \). For a voxel query \(\mathbf{Q}\) at \((i, j, k)\), it interacts only with nearby reference points. The voxel self-attention process is formulated as:

\begin{equation}
    \text{TSA}(\mathbf{Q},\mathbf{Q_{k}})= \sum_{n=1}^{N_2}\text{DA}(\mathbf{Q},\mathbf{Ref}_{i,j,k}^{n},\mathbf{Q_{k}})
\end{equation}

where $n$ indexes reference points, $N_2$ is the total number of reference points per voxel query. $\text{DA}$ denotes deformable attention. $\mathbf{Ref}_{i,j,k}^{n}$ in voxel self-attention is defined on the BEV plane, unlike image-plane reference points in voxel cross-attention.
\begin{equation}
    \mathbf{Ref}_{i,j,k}^{n} = (x_i^n,y_j^n,z_k)
\end{equation}
where $(x_i^n, y_j^n, z_k)$ indexes the $n$-th reference point for query $\mathbf{Q}$, sharing height $z_k$ but with different learnable offsets for $(x_i^n, y_j^n)$. 

We embed voxel features aggregated from adjacent timestamps as local temporal conditions into the denoising network, obtaining fine-grained perception information from these moments.

\noindent\textbf{Global Temporal Interaction.}
Voxel features capture fine geometric details crucial for 3D structures, but maintaining global perception using voxel representations is impractical due to storage and computational limitations\cite{ye2021drinet++,tang2020searching,sitzmann2019deepvoxels}.
BEV representations\cite{li2022bevstereo,huang2021bevdet,li2022bevdepth}provide a clear global view of object positions and scales [16, 10, 14], address occlusion, and are more memory- and computation-efficient. Therefore, we use BEV as the global representation for temporal queue features.

Drawing on previous work\cite{xia2024henethybridencodingendtoend,Liu2024LetOF}, we compress the locally aligned feature volume \( V_t \) into a series of BEV slices \( B_t \in \mathbb{R}^{H \times W \times C} \) along the \( Z \)-axis and apply deformable attention (DA) horizontally to interact among these BEV slices. To efficiently extract overall information from the vertical dimension, we further generate a sequence of globally aligned BEV feature maps through average pooling operations\cite{FBBEV,BEV-LaneDet,bevpoolv2,fastBEV}.

\begin{figure}[htbp]
    \centering
    \includegraphics[width=0.5\textwidth, keepaspectratio]{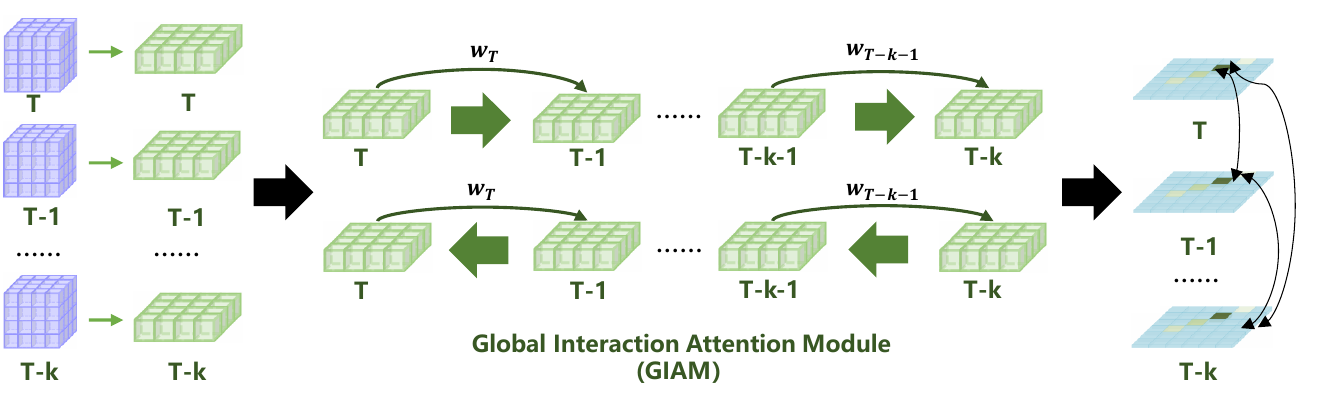}  
    \caption{The Global Temporal Information Interaction Module consists of Voxel Feature Mapping and Global Interaction Attention Module. Voxel Feature Mapping pools time-aligned voxels into BEV space to retain global information. Global Interaction Attention Module employs deformable attention with time-decay weights and achieves temporal interaction through a backward-forward process.}  
    \label{fig:GIAM}  
\end{figure}

Next, we employ a global temporal interaction method to enable interactions among temporal BEV features within the temporal queue. This temporal fusion process consists of two components: a backward process, which fuses current frame features into past frames, and a forward process, which aggregates past frame features into the current frame.In each step, a cross-attention-based adjacent frame fusion module, named the Global Interaction Attention Module (GIAM), is used to fuse BEV features from adjacent frames.
Specifically, for BEV features $B_{t_i}$ and $B_{t_j}$, the progress of interaction can be formulated as:

\begin{equation}
\mathbf{GIAM}(B_{t_i}, B_{t_j}) = B_{t_j} + \omega(t) \times G
\end{equation}

\begin{equation}
\text{G}=\text{Avg}\left( 
     \text{DA}(\langle B_{t_i}, B_{t_j} \rangle, B_{t_i}),
 \text{DA}(\langle B_{t_i}, B_{t_j} \rangle, B_{t_j})\right)
\end{equation}

where G denotes the information gain from features of adjacent timestamps. $\omega$ is is a learnable weight that decays over time. $Avg(\cdot)$ represents the average operator, $\langle \cdot, \cdot \rangle$ denotes concatenation.
In the backward process, $t_j=t_i-1$, while in the forward process, $t_j=t_i+1$.

By constructing the feature sequence this way, we supply the denoising network with features for global temporal aggregation while reducing computational costs.

\subsection{In-model Latent Denoising Network}
\noindent\textbf{Input Representation.} As shown in Figure~\ref{fig:pipeline}, Gaussian noise is added to the coarse BEV feature map to obtain the noisy BEV feature map at the current timestamp. The noisy embedding \( \mathbf{x}_i \) without temporal information is used as input, conditioned on locally aggregated voxel feature and globally aggregated historical BEV features. Meanwhile, the local and global temporal features are mapped to embeddings t.

\noindent\textbf{Denoising Block.} The inputs to the denoising block\cite{zhou2025detrackinmodellatentdenoising} include the noisy BEV feature map embedding and the temporal condition embedding as shown in Figure~\ref{fig:pipeline}. These are passed through linear layers to obtain \( \mathbf{q}_{x_i} \) (current BEV feature), \( \mathbf{k}_t \) (temporal key), and \( \mathbf{v}_t \) (temporal value).

Subsequently, the denoising attention mechanism is first employed for denoising:

\begin{equation}
    \text{Attention}_{Denoising}(t, \mathbf{x}_i) = \text{Softmax}(\frac{q_{\mathbf{x}_i}k_{t}}{\sqrt{d}}v_{t}). 
\end{equation}

Then, we introduce a Feedforward Neural Network (FFN) layer to enhance \( \mathbf{x}_i \):

\begin{equation}
        \mathbf{x}^{'}_i =\text{Attention}_{Denoising}(t, \mathbf{x}_i)+\mathbf{x}_i.
\end{equation}
\vspace{-2mm}
\begin{equation}
        \mathbf{x}^{''}_i = \mathbf{x}^{'}_i+\text{FFN}(\mathbf{x}^{'}_i),
\end{equation}

Finally, we use two linear layers to predict the noise for the second denoising step. The noise is subtracted from the embedding to obtain the denoised result after the noise prediction module:

\begin{equation}
\begin{aligned}
    \epsilon = \text{NoisePred}(\mathbf{x}^{''}_i)&=\text{Linear}(\text{ReLu}(\text{Linear}(\mathbf{x}^{''}_i))),\\
    \mathbf{x}_{i-\frac{I}{l}} &= \mathbf{x}^{''}_i-\epsilon.
\end{aligned}
\end{equation}
\vspace{-2mm}

Denoising is performed through \( l \) denoising blocks. Ultimately, denoising is completed with a single forward pass of the in-model latent denoising network, resulting in the denoised BEV feature embedding.
\begin{equation}
    \mathbf{x}_0 =\mathbf{x}_I-\sum_{j=1}^{l}\epsilon_j.
\end{equation}
We concatenate the obtained BEV feature maps to aggregate voxel features with global temporal information.

\definecolor{nbarrier}{RGB}{255, 120, 50}
\definecolor{nbicycle}{RGB}{255, 192, 203}
\definecolor{nbus}{RGB}{255, 255, 0}
\definecolor{ncar}{RGB}{0, 150, 245}
\definecolor{nconstruct}{RGB}{0, 255, 255}
\definecolor{nmotor}{RGB}{200, 180, 0}
\definecolor{npedestrian}{RGB}{255, 0, 0}
\definecolor{ntraffic}{RGB}{255, 240, 150}
\definecolor{ntrailer}{RGB}{135, 60, 0}
\definecolor{ntruck}{RGB}{160, 32, 240}
\definecolor{ndriveable}{RGB}{255, 0, 255}
\definecolor{nother}{RGB}{139, 137, 137}
\definecolor{nsidewalk}{RGB}{75, 0, 75}
\definecolor{nterrain}{RGB}{150, 240, 80}
\definecolor{nmanmade}{RGB}{213, 213, 213}
\definecolor{nvegetation}{RGB}{0, 175, 0}
\definecolor{nothers}{RGB}{0, 0, 0}
\begin{table*}[ht]
	\footnotesize
 	\setlength{\tabcolsep}{0.0025\linewidth}
	
	\newcommand{\classfreq}[1]{{~\tiny(\nuscenesfreq{#1}\%)}}  %
    \begin{center}

	\begin{tabular}{l|c|c|c| c c c c c c c c c c c c c c c c c}
		\toprule
		Method

		& Modality
            & Backbone
            & mIoU
        
        & \rotatebox{90}{\textcolor{nothers}{$\blacksquare$} others}
        
		& \rotatebox{90}{\textcolor{nbarrier}{$\blacksquare$} barrier}
		
		& \rotatebox{90}{\textcolor{nbicycle}{$\blacksquare$} bicycle}
		
		& \rotatebox{90}{\textcolor{nbus}{$\blacksquare$} bus}

		& \rotatebox{90}{\textcolor{ncar}{$\blacksquare$} car}

		& \rotatebox{90}{\textcolor{nconstruct}{$\blacksquare$} const. veh.}

		& \rotatebox{90}{\textcolor{nmotor}{$\blacksquare$} motorcycle}

		& \rotatebox{90}{\textcolor{npedestrian}{$\blacksquare$} pedestrian}

		& \rotatebox{90}{\textcolor{ntraffic}{$\blacksquare$} traffic cone}

		& \rotatebox{90}{\textcolor{ntrailer}{$\blacksquare$} trailer}

		& \rotatebox{90}{\textcolor{ntruck}{$\blacksquare$} truck}

		& \rotatebox{90}{\textcolor{ndriveable}{$\blacksquare$} drive. suf.}

		& \rotatebox{90}{\textcolor{nother}{$\blacksquare$} other flat}

		& \rotatebox{90}{\textcolor{nsidewalk}{$\blacksquare$} sidewalk}

		& \rotatebox{90}{\textcolor{nterrain}{$\blacksquare$} terrain}

		& \rotatebox{90}{\textcolor{nmanmade}{$\blacksquare$} manmade}

		& \rotatebox{90}{\textcolor{nvegetation}{$\blacksquare$} vegetation}

		\\
		\midrule
        MonoScene~\cite{cao2022monoscene} & Camera & R101-DCN & 6.06 & 1.75 & 7.23 & 4.26 & 4.93 & 9.38 & 5.67 & 3.98 & 3.01 & 5.90 & 4.45 & 7.17 & 14.91 & 6.32 & 7.92 & 7.43 & 1.01 & 7.65\\
        BEVDet~\cite{huang2021bevdet} & Camera & R101-DCN & 11.73 & 2.09 & 15.29 & 0.0 & 4.18 & 12.97 & 1.35 & 0.0 & 0.43 & 0.13 & 6.59 & 6.66 & 52.72 & 19.04 & 26.45 & 21.78 & 14.51 & 15.26\\
		BEVFormer~\cite{BEVFormer} & Camera & R101-DCN & 26.88 & 5.85 & 37.83 & 17.87 & 40.44 & 42.43 & 7.36 & 23.88 & 21.81 & 20.98 & 22.38 & 30.70 & 55.35 & 28.36 & 36.0 & 28.06 & 20.04 & 17.69 \\
        CTF-Occ~\cite{tian2023occ3d} & Camera & R101-DCN & 28.53 & 8.09 & 39.33 & 20.56 & 38.29 & 42.24 & 16.93 & 24.52 & 22.72 & 21.05 & 22.98 & 31.11 & 53.33 & 33.84 & 37.98 & 33.23 & 20.79 & 18.0\\
        \midrule
        \  PanoOcc\(*\) & Camera & R101-DCN & 36.63 & 8.64 & 43.71 & 21.69 & 42.55 & 49.91 & 21.32 & 25.35 & 22.92 & 20.19 & 29.78 & 37.19 & 80.97 & 40.36 & 49.65 & 52.84 & 39.81 & 35.82  \\
        \  PanoOcc & Camera & R101-DCN & 42.13 & 11.67 & 50.48 & 29.64 & 49.44 & 55.52 & 23.29 & 33.26 & 30.55 & 30.99 & 34.43 & 42.57 & 83.31 & 44.23 & 54.4 & 56.04 & 45.94 & 40.4  \\
        \  \textbf{GTAD\(*\)} &  Camera & R101-DCN & \textbf{40.76} & 6.42 & 43.40 & \textbf{22.11} & \textbf{52.62} & \textbf{56.64} & \textbf{29.79} & 23.51 & \textbf{26.79} & \textbf{20.45} & \textbf{41.81} & \textbf{46.60} & \textbf{82.91} & \textbf{45.76} & \textbf{51.80} & \textbf{60.05} & \textbf{42.96} & \textbf{38.61}  \\
		\bottomrule
	\end{tabular}
    \end{center}
    \caption{\textbf{3D Occupancy Prediction performance on the Occ3D-nuScenes dataset.}
    \(*\) represents that the model was trained for 12 epochs.}
 \vspace{-10pt}
	\label{tab:occ}
\end{table*}
\definecolor{nbarrier}{RGB}{255, 120, 50}
\definecolor{nbicycle}{RGB}{255, 192, 203}
\definecolor{nbus}{RGB}{255, 255, 0}
\definecolor{ncar}{RGB}{0, 150, 245}
\definecolor{nconstruct}{RGB}{0, 255, 255}
\definecolor{nmotor}{RGB}{200, 180, 0}
\definecolor{npedestrian}{RGB}{255, 0, 0}
\definecolor{ntraffic}{RGB}{255, 240, 150}
\definecolor{ntrailer}{RGB}{135, 60, 0}
\definecolor{ntruck}{RGB}{160, 32, 240}
\definecolor{ndriveable}{RGB}{255, 0, 255}
\definecolor{nother}{RGB}{139, 137, 137}
\definecolor{nsidewalk}{RGB}{75, 0, 75}
\definecolor{nterrain}{RGB}{150, 240, 80}
\definecolor{nmanmade}{RGB}{213, 213, 213}
\definecolor{nvegetation}{RGB}{0, 175, 0}
\begin{table*}[ht]
	\footnotesize
 	\setlength{\tabcolsep}{0.0045\linewidth}
	
	\newcommand{\classfreq}[1]{{~\tiny(\nuscenesfreq{#1}\%)}}  %
    \begin{center}

	\begin{tabular}{l|c|c|c | c c c c c c c c c c c c c c c c}
		\toprule
		Method

		& Modality
            & Backbone
            & mIoU
        
		& \rotatebox{90}{\textcolor{nbarrier}{$\blacksquare$} barrier}
		
		& \rotatebox{90}{\textcolor{nbicycle}{$\blacksquare$} bicycle}
		
		& \rotatebox{90}{\textcolor{nbus}{$\blacksquare$} bus}

		& \rotatebox{90}{\textcolor{ncar}{$\blacksquare$} car}

		& \rotatebox{90}{\textcolor{nconstruct}{$\blacksquare$} const. veh.}

		& \rotatebox{90}{\textcolor{nmotor}{$\blacksquare$} motorcycle}

		& \rotatebox{90}{\textcolor{npedestrian}{$\blacksquare$} pedestrian}

		& \rotatebox{90}{\textcolor{ntraffic}{$\blacksquare$} traffic cone}

		& \rotatebox{90}{\textcolor{ntrailer}{$\blacksquare$} trailer}

		& \rotatebox{90}{\textcolor{ntruck}{$\blacksquare$} truck}

		& \rotatebox{90}{\textcolor{ndriveable}{$\blacksquare$} drive. suf.}

		& \rotatebox{90}{\textcolor{nother}{$\blacksquare$} other flat}

		& \rotatebox{90}{\textcolor{nsidewalk}{$\blacksquare$} sidewalk}

		& \rotatebox{90}{\textcolor{nterrain}{$\blacksquare$} terrain}

		& \rotatebox{90}{\textcolor{nmanmade}{$\blacksquare$} manmade}

		& \rotatebox{90}{\textcolor{nvegetation}{$\blacksquare$} vegetation}

		\\
		\midrule

        BEVFormer-Base~\cite{BEVFormer} & Camera &R101-DCN & 56.2  & 54.0 & 22.8 & 76.7 & 74.0 & 45.8 & 53.1 & 44.5 & 24.7 & 54.7 & 65.5 & 88.5 & 58.1 & 50.5 & 52.8 & 71.0 & 63.0  \\
		
		TPVFormer-Base~\cite{huang2023tri}  & Camera & R101-DCN &68.9  & 70.0 & 40.9 & 93.7 & 85.6 & 49.8 & 68.4 & 59.7 & 38.2 & 65.3 & 83.0 & 93.3 & 64.4 & 64.3 & 64.5 & 81.6 & 79.3  \\ %
         \midrule

        \ PanoOcc-Small-T &  Camera & R50 &68.1 & 70.7 & 37.9 & 92.3 & 85.0 & 50.7 & 64.3 & 59.4 & 35.3 & 63.8 & 81.6 & 94.2 & 66.4 & 64.8 & 68.0 & 79.1 & 75.6 \\

         \ \textbf{GTAD} &  Camera & R50 &\textbf{70.2}  & \textbf{71.2} & 36.5 & 93.1 & 84.7 & \textbf{51.3} & 65.2 & \textbf{60.1} & 34.9 & 64.5 & 82.3 & \textbf{95.0} & \textbf{67.1} & \textbf{65.6} & \textbf{69.8} & 78.9 & 74.4 \\ %
  
		\bottomrule
	\end{tabular}
    \end{center}
    \caption{\textbf{LiDAR Semantic Segmentation results on nuScenes validation set.} 
	}
 \vspace{-10pt}
	\label{tab:lidar_seg}
\end{table*}


\begin{figure*}[htbp]
    \centering
    \includegraphics[width=\textwidth]
    {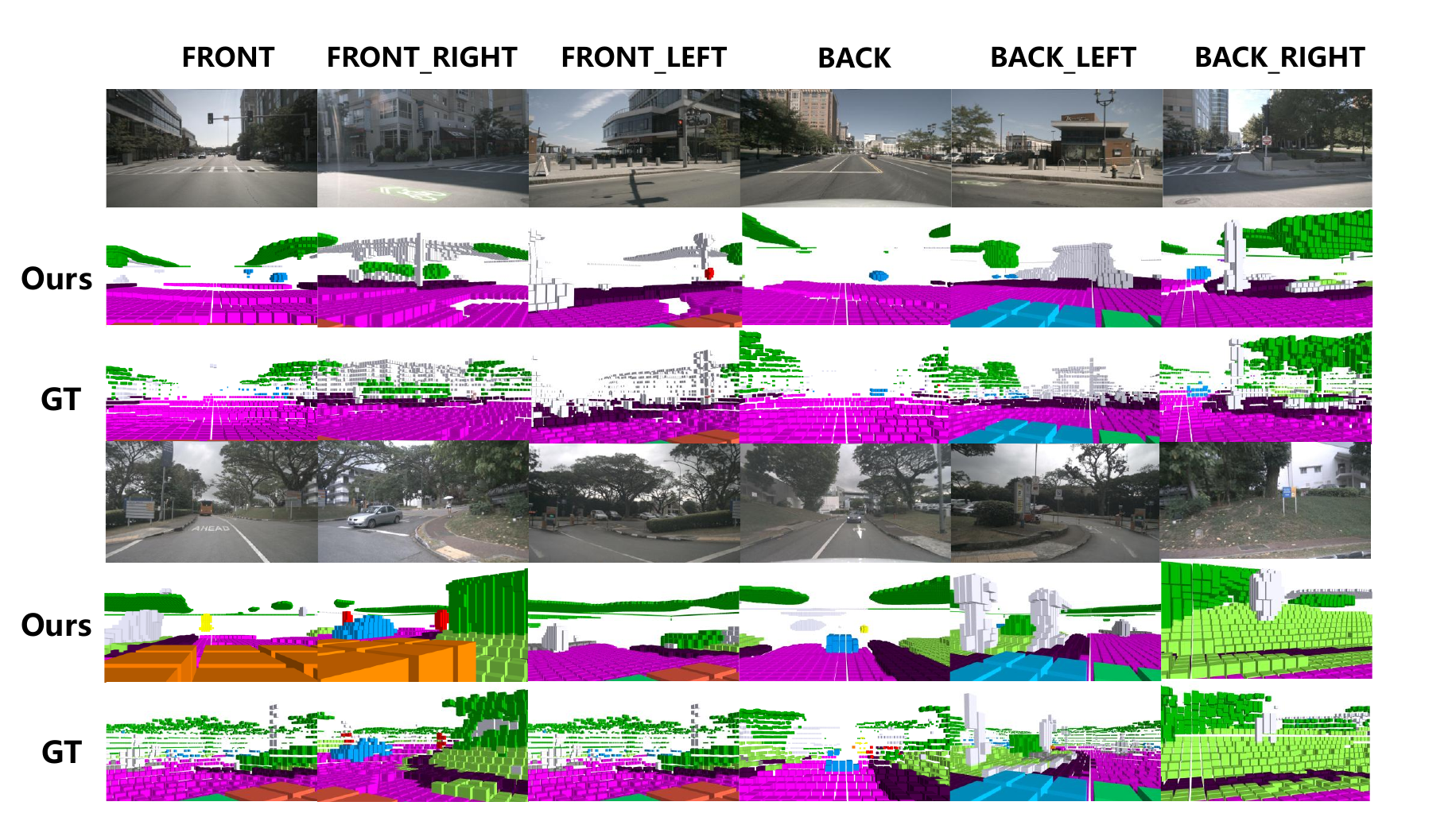}  
    \caption{\textbf{Our method is compared with the ground truth of Occ3d-nuScenes through visualizations in various scenarios.} In addition to urban areas, we selected challenging scenarios. The results demonstrate that our method performs well across different scenarios.}  
    \label{fig:display}  
\end{figure*}
\subsection{Loss Function}
PanoOcc unifies 3D object detection and semantic segmentation into 3D panoptic segmentation by jointly training them with foreground information propagation. This approach enables comprehensive scene understanding and fine-grained object modeling. Referring to this method our model is end-to-end trained for joint detection and segmentation with a total loss \( L \) composed of \( L_{\text{Det}} \) and \( L_{\text{Seg}} \).

The semantic voxel segmentation head is supervised by \( L_{\text{Seg}} \), which includes focal loss\cite{lin2017focal}(applied to all voxels) and Lovász loss\cite{berman2018lovasz} (applied to non-empty voxels). Foreground information is propagated to the detection head, predicting a binary voxel mask for foreground classes (things), supervised by focal loss \( L_{\text{thing}} \).

The total segmentation loss \( L_{\text{Seg}} \) is formulated as:

\begin{equation}
L_{Seg} = \lambda_1 L_{\text{focal}} + \lambda_2 L_{\text{lovasz}} + \lambda_3 L_{\text{thing}} \quad 
\end{equation}

The detection head is supervised by $L_{\text{Det}}$, a sparse loss consisting of focal loss\cite{lin2017focal} for classification and $L1$ loss for bounding box regression:

\begin{equation}
L_{\text{Det}} = \lambda_4 L_{\text{cls}} + \lambda_5 L_{\text{reg}} \quad 
\end{equation}
By leveraging the advantages of a joint task training framework, our GTAD achieves a more comprehensive understanding of the scene.

\begin{table*}[ht]
	\footnotesize
 	\setlength{\tabcolsep}{0.011\linewidth}

	\newcommand{\classfreq}[1]{{~\tiny(\nuscenesfreq{#1}\%)}}  %
    \begin{center}

	\begin{tabular}{l|c| c c c c c c c c c c c c c c c c}
		\toprule
		\makecell[c]{using\\ local temporal\\aggregation}

            & mIoU
        
		& \rotatebox{90}{barrier}
		
		& \rotatebox{90}{bicycle}
		
		& \rotatebox{90}{bus}

		& \rotatebox{90}{car}

		& \rotatebox{90}{const. veh.}

		& \rotatebox{90}{motorcycle}

		& \rotatebox{90}{pedestrian}

		& \rotatebox{90}{traffic cone}

		& \rotatebox{90}{trailer}

		& \rotatebox{90}{truck}

		& \rotatebox{90}{drive. suf.}

		& \rotatebox{90}{other flat}

		& \rotatebox{90}{sidewalk}

		& \rotatebox{90}{terrain}

		& \rotatebox{90}{manmade}

		& \rotatebox{90}{vegetation}

		\\
		\midrule
        \makecell[c]{$\times$}  &66.1  & 62.0 & 35.5 & 91.2 & 83.1 & 48.7 & 65.3 & 58.5 & 33.8 & 64.0 & 82.5 & 93.5 & 63.4 & 63.3 & 63.6 & 77.7 & 72.4\\
         \makecell[c]{$\checkmark$}  &70.2  & 71.2 & 36.5 & 93.1 & 84.7 & 51.3 & 65.2 & 60.1 & 34.9 & 64.5 & 82.3 & 95.0 & 67.1 & 65.6 & 69.8 & 78.9 & 74.4\\
	\makecell[c]{$\uparrow$} & \textbf{4.1} & \textbf{9.2} & \textbf{1.0} & \textbf{1.9} & \textbf{1.6} & \textbf{2.6} & -0.1 & \textbf{1.6} & \textbf{1.1} & \textbf{0.5} & -0.2 & \textbf{1.5} & \textbf{3.7} & \textbf{2.3} & \textbf{6.2} & \textbf{1.2} & \textbf{2.0} \\

		\bottomrule
	\end{tabular}
    \end{center}
    \caption{\textbf{Ablation study of local temporal encoder for LiDAR segmentation.} }
 \vspace{-10pt}
	\label{tab:use local}
\end{table*}

\begin{table*}[ht]
	\footnotesize
 	\setlength{\tabcolsep}{0.031\linewidth}

	\newcommand{\classfreq}[1]{{~\tiny(\nuscenesfreq{#1}\%)}}  %
    \begin{center}

	\begin{tabular}{l|c c c c c c c c c}
		\toprule
            & step1
        
		& step2
		
		&step3
		
		&step4

		& step5

		& \textbf{step6}
        
		& step7

		& step8

		& step9
		\\
		\midrule
        mIoU &4.1 &12.6 &24.8&26.5 &37.54 &\textbf{40.76}&36.1&37.3&35.7\\
		\bottomrule
	\end{tabular}
    \end{center}
    \caption{\textbf{Ablation study of denoising steps. The best results are highlighted in bold.} }
 \vspace{-10pt}
    \label{tab:noise step}
\end{table*}

\section{Experiments}
\subsection{Datasets}

\noindent\textbf{nuScenes dataset\cite{caesar2020nuscenes}} contains 1,000 scenes, which are divided into 700 training scenes, 150 validation scenes, and 150 testing scenes. Each scene is recorded for 20 seconds at a sampling rate of 20 Hz. Each sample includes RGB images from six cameras, covering a 360$^{\circ}$ horizontal field of view, as well as point cloud data generated by a 32-beam lidar sensor. For the object detection task, key samples are annotated at a frequency of 2 Hz, covering ground-truth labels for 10 foreground object classes. For the semantic segmentation tasks, each lidar point in the key samples is labeled with one of 6 background classes and 10 foreground classes.

\noindent\textbf{Occ3D-nuScenes\cite{tian2023occ3d}} 
contains 700 training scenes and 150 validation scenes. The occupancy range is defined as 40 meters to 40 meters along the X and Y axes, and 1 meter to 5.4 meters along the Z axis. The voxel size is \(0.4 \, \text{meters} \times 0.4 \, \text{meters} \times 0.4 \, \text{meters}\). The semantic labels include 17 classes (including “others”).

\subsection{Experimental Settings}
\noindent\textbf{Input Settings.}
On the nuScenes dataset\cite{caesar2020nuscenes}, we set the range of the point cloud for the \(x\) and \(y\) axes to \([-51.2\, \text{m}, 51.2\, \text{m}]\) and the \(z\) axis to \([-5\, \text{m}, 3\, \text{m}]\). The voxel grid size used for loss supervision is \((0.256\, \text{m}, 0.256\, \text{m}, 0.125\, \text{m})\). The input image size is cropped to \(640 \times 1600\). When using the R101-DCN \cite{wang2022internimage} as the backbone network, the default image scale is \(1.0\) (\(640 \times 1600\)). When using the R50\cite{he2016deep} backbone network, the image scale is set to \(0.5\) (\(320 \times 800\)).The initial resolution of the voxel query is \(50 \times 50 \times 16\), corresponding to \(H\), \(W\), and \(Z\), respectively. We use an embedding dimension of 128 and add learnable 3D positional encodings to the voxel queries.

\noindent\textbf{Occupancy Decoder and Task Head.} 
The voxel upsampling module uses a 3-layer 3D transposed convolution to upsample \(H\) and \(W\) by a factor of 4 and \(Z\) by a factor of 2. The upsampled voxel features have a resolution of \(200 \times 200 \times 32\) with a feature dimension \(D_0\) of 64.The segmentation head consists of a 2-layer MLP with a hidden dimension of 128, using the softplus\cite{zheng2015improving} activation function. The detection head is configured with 900 object queries and a 6-layer decoder, similar to \cite{BEVFormer}.

\subsection{Train and Evalution}
\noindent\textbf{Training.}   We train the model on 8 NVIDIA RTX4090 GPUs with a batch size of 1 per GPU. Throughout the training process, we use the AdamW optimizer\cite{loshchilov2017decoupled} for 12 epochs. For the detection head, we use object-level annotations as supervision. We supervise the voxel prediction of the segmentation head with sparse LiDAR point-level semantic labels.  For occupancy prediction, we rely on occupancy labels as the source of supervision.

\noindent\textbf{3D Occupancy Prediction.}   
For the dense evaluation of the occupancy benchmark\cite{tian2023occ3d}, we directly compute the mIoU based on occupancy labels.

\noindent\textbf{LiDAR Semantic Segmentation.}   
We use R101-DCN initialized from the FCOS3D\cite{wang2021fcos3d} checkpoint. This is consistent with TPVFormer\cite{huang2023tri} and BEVFormer\cite{BEVFormer}.And we also compute the mIoU based on LiDAR point-level semantic labels.

\subsection{Main Results}

\noindent\textbf{3D Occupancy Prediction.}
In Table~\ref{tab:occ}, we evaluate our method for 3D occupancy prediction on the Occ3D-nuScenes validation set\cite{tian2023occ3d}. All compared methods use camera inputs. Our method is trained for 12 epochs, while PanoOcc is trained for both 12 and 24 epochs. Other methods, including MonoScene\cite{cao2022monoscene}, BEVDet\cite{huang2021bevdet}, BEVFormer\cite{BEVFormer}, and CTF Occ\cite{tian2023occ3d}, are trained for 24 epochs, with their performance reported in \cite{tian2023occ3d}. We employ an R101-DCN backbone network and perform temporal fusion using 4 frames. Figure~\ref{fig:display} illustrates the dense occupancy prediction on the Occ3D-nuScenes validation set.In comparison with others, we have got a good performance.

Compared to PanoOcc, our method achieves higher mIoU at 12 epochs but falls short of its 24-epoch version. However, in terms of memory consumption, our approach uses significantly less memory than PanoOcc trained for 24 epochs, as shown in Table~\ref{tab:gpu with pano}.

\noindent\textbf{LiDAR Semantic Segmentation.}
We evaluate our model's performance on the nuScenes test and validation sets. In Table~\ref{tab:lidar_seg}, we conduct experiments using two different backbone networks. Under the R50 setting, our method achieves a 70.2 mIoU, outperforming other camera-based detection methods and showing improvement compared to the small-T version of PanoOcc.

\begin{table}[t]
    \small
    \begin{center}
    \setlength{\tabcolsep}{20pt}
    \begin{tabular}{c|c|c}
		\toprule
		& Fuse  & mIoU \\
		\midrule
      (a)  & Cat   & 34.25 \\
      (b)   & TSA    & 30.13 \\
      (c)   & GIAM    & \textbf{40.76}\\
		\bottomrule
	\end{tabular}
    \end{center}
     \caption{\textbf{Ablation study for Global temporal encoder.}
     }
    \label{tab:fuse}
\end{table}

\subsection{Ablation Study}
\noindent\textbf{Design of the Global Temporal Encoder.}
Table~\ref{tab:fuse} presents an ablation study on global encoder design. Direct feature concatenation in the temporal queue outperforms temporal self-attention\cite{BEVFormer}, and the global temporal interaction module significantly improves performance.

\noindent\textbf{Local Temporal Encoder.}    
Table~\ref{tab:use local} shows the impact of aggregating local temporal information at current time on different classes. Most classes saw improved semantic segmentation performance.
\begin{table}[t]
    \small
    \begin{center}
    \setlength{\tabcolsep}{20pt}
    \begin{tabular}{c|c}
		\toprule
		timestep & mIoU \\
		\midrule
        200   & 6.70 \\ 
         400   & 20.02 \\
          600   & 32.31 \\
            \textbf{800}   & \textbf{40.76}\\ 
           1000  & 38.42\\ 
        
		\bottomrule
	\end{tabular}
    \end{center}
     \caption{\textbf{Numeric ablation
 studies on different corruption
 scales in Denoising  Network} }
    \label{tab:timestep}
    \vspace{-10pt}
\end{table}

\begin{table}[t]
    \small
    \begin{center}
    \setlength{\tabcolsep}{20pt}
    \begin{tabular}{c|c|c}
		\toprule
		Epoch&Method & Memory (MB) \\
		\midrule
        12&GTAD& \textbf{19901} \\
        12&PanoOcc& 14400 \\
        24&PanoOcc& \textbf{35400} \\
        
		\bottomrule
	\end{tabular}
    \end{center}
     \caption{\textbf{Compare our method with PanoOcc by Memory} }
    \label{tab:gpu with pano}
    \vspace{-10pt}
\end{table}

\noindent\textbf{Influence of Denoising Steps.}   
We studied the impact of denoising iterations on performance. Our model has 6 denoising steps using forward-only blocks. Performance is not good at step 1, but improves significantly by step 3, and peaks at step 6 (Table~\ref{tab:noise step}).

\noindent\textbf{Numeric Ablation on Corruption Scale.}   
Table~\ref{tab:timestep} lists the mIoU values at different corruption scales \(t\). It can be seen that the best performance is achieved when \(t = 800\). A higher number of time steps helps the model find the global optimum and fully aggregate temporal conditions,however, an excessively large time step can lead to a decline in performance.

\section{Conclusion}
In this paper, we introduce GTAD, a novel denoising-based framework for 3D semantic occupancy prediction. GTAD leverages an in-model latent denoising network to aggregate both local and global temporal features, effectively capturing fine-grained and historical temporal information. Additionally, an attention-based feature interaction mechanism is employed to integrate global historical context, thereby addressing the limitations of prior approaches. This results in voxel features that facilitate a more coherent and comprehensive understanding of the scene. Extensive experiments on the nuScenes and Occ3D-nuScenes benchmark and ablation studies demonstrate the  superiority of our method.
We foresee that the efficient exploitation of temporal information represents a promising new paradigm for camera-based 3D scene perception.

\bibliographystyle{IEEEtran}
\bibliography{IEEEabrv,mybibtex}

\begin{thebibliography}{10}
\providecommand{\url}[1]{#1}
\csname url@rmstyle\endcsname
\providecommand{\newblock}{\relax}
\providecommand{\bibinfo}[2]{#2}
\providecommand\BIBentrySTDinterwordspacing{\spaceskip=0pt\relax}
\providecommand\BIBentryALTinterwordstretchfactor{4}
\providecommand\BIBentryALTinterwordspacing{\spaceskip=\fontdimen2\font plus
\BIBentryALTinterwordstretchfactor\fontdimen3\font minus \fontdimen4\font\relax}
\providecommand\BIBforeignlanguage[2]{{%
\expandafter\ifx\csname l@#1\endcsname\relax
\typeout{** WARNING: IEEEtran.bst: No hyphenation pattern has been}%
\typeout{** loaded for the language `#1'. Using the pattern for}%
\typeout{** the default language instead.}%
\else
\language=\csname l@#1\endcsname
\fi
#2}}

\bibitem{zhou2022cross}
B.~Zhou and P.~Kr{\"a}henb{\"u}hl, ``Cross-view transformers for real-time map-view semantic segmentation,'' in \emph{Proceedings of the IEEE/CVF Conference on Computer Vision and Pattern Recognition}, 2022, pp. 13\,760--13\,769.

\bibitem{milioto2020lidar}
A.~Milioto, J.~Behley, C.~McCool, and C.~Stachniss, ``Lidar panoptic segmentation for autonomous driving,'' in \emph{2020 IEEE/RSJ International Conference on Intelligent Robots and Systems (IROS)}.\hskip 1em plus 0.5em minus 0.4em\relax IEEE, 2020, pp. 8505--8512.

\bibitem{hu2021fiery}
A.~Hu, Z.~Murez, N.~Mohan, S.~Dudas, J.~Hawke, V.~Badrinarayanan, R.~Cipolla, and A.~Kendall, ``Fiery: future instance prediction in bird's-eye view from surround monocular cameras,'' in \emph{Proceedings of the IEEE/CVF International Conference on Computer Vision}, 2021, pp. 15\,273--15\,282.

\bibitem{dd3d}
D.~Park, R.~Ambrus, V.~Guizilini, J.~Li, and A.~Gaidon, ``Is pseudo-lidar needed for monocular 3d object detection?'' in \emph{ICCV}, 2021.

\bibitem{mmdet3d2020}
M.~Contributors, ``{MMDetection3D: OpenMMLab} next-generation platform for general {3D} object detection,'' \url{https://github.com/open-mmlab/mmdetection3d}, 2020.

\bibitem{liao2022maptr}
B.~Liao, S.~Chen, X.~Wang, T.~Cheng, Q.~Zhang, W.~Liu, and C.~Huang, ``Maptr: Structured modeling and learning for online vectorized hd map construction,'' \emph{arXiv preprint arXiv:2208.14437}, 2022.

\bibitem{roddick2020predicting}
T.~Roddick and R.~Cipolla, ``Predicting semantic map representations from images using pyramid occupancy networks,'' in \emph{CVPR}, 2020.

\bibitem{liu2022spatial}
J.~Liu, Y.~Chen, X.~Ye, Z.~Tian, X.~Tan, and X.~Qi, ``Spatial pruned sparse convolution for efficient 3d object detection,'' \emph{arXiv preprint arXiv:2209.14201}, 2022.

\bibitem{li2021multi}
S.~Li, X.~Chen, Y.~Liu, D.~Dai, C.~Stachniss, and J.~Gall, ``Multi-scale interaction for real-time lidar data segmentation on an embedded platform,'' \emph{IEEE Robotics and Automation Letters}, vol.~7, no.~2, pp. 738--745, 2021.

\bibitem{miao2023occdepth}
R.~Miao, W.~Liu, M.~Chen, Z.~Gong, W.~Xu, C.~Hu, and S.~Zhou, ``Occdepth: A depth-aware method for 3d semantic scene completion,'' \emph{arXiv preprint arXiv:2302.13540}, 2023.

\bibitem{huang2023tri}
Y.~Huang, W.~Zheng, Y.~Zhang, J.~Zhou, and J.~Lu, ``Tri-perspective view for vision-based 3d semantic occupancy prediction,'' \emph{arXiv preprint arXiv:2302.07817}, 2023.

\bibitem{wang2023openoccupancy}
X.~Wang, Z.~Zhu, W.~Xu, Y.~Zhang, Y.~Wei, X.~Chi, Y.~Ye, D.~Du, J.~Lu, and X.~Wang, ``Openoccupancy: A large scale benchmark for surrounding semantic occupancy perception,'' \emph{arXiv preprint arXiv:2303.03991}, 2023.

\bibitem{li2023voxformer}
Y.~Li, Z.~Yu, C.~Choy, C.~Xiao, J.~M. Alvarez, S.~Fidler, C.~Feng, and A.~Anandkumar, ``Voxformer: Sparse voxel transformer for camera-based 3d semantic scene completion,'' in \emph{Proceedings of the IEEE/CVF Conference on Computer Vision and Pattern Recognition}, 2023, pp. 9087--9098.

\bibitem{BEVFormer}
Z.~Li, W.~Wang, H.~Li, E.~Xie, C.~Sima, T.~Lu, Y.~Qiao, and J.~Dai, ``Bevformer: Learning bird’s-eye-view representation from multi-camera images via spatiotemporal transformers,'' in \emph{ECCV}, 2022.

\bibitem{Liu2024LetOF}
\BIBentryALTinterwordspacing
Y.~Liu, L.~Mou, X.~Yu, C.~Han, S.~Mao, R.~Xiong, and Y.~Wang, ``Let occ flow: Self-supervised 3d occupancy flow prediction,'' \emph{ArXiv}, vol. abs/2407.07587, 2024. [Online]. Available: \url{https://api.semanticscholar.org/CorpusID:271088768}
\BIBentrySTDinterwordspacing

\bibitem{wang2023panooccunifiedoccupancyrepresentation}
\BIBentryALTinterwordspacing
Y.~Wang, Y.~Chen, X.~Liao, L.~Fan, and Z.~Zhang, ``Panoocc: Unified occupancy representation for camera-based 3d panoptic segmentation,'' 2023. [Online]. Available: \url{https://arxiv.org/abs/2306.10013}
\BIBentrySTDinterwordspacing

\bibitem{xia2024henethybridencodingendtoend}
\BIBentryALTinterwordspacing
Z.~Xia, Z.~Lin, X.~Wang, Y.~Wang, Y.~Xing, S.~Qi, N.~Dong, and M.-H. Yang, ``Henet: Hybrid encoding for end-to-end multi-task 3d perception from multi-view cameras,'' 2024. [Online]. Available: \url{https://arxiv.org/abs/2404.02517}
\BIBentrySTDinterwordspacing

\bibitem{huang2022bevdet4dexploittemporalcues}
\BIBentryALTinterwordspacing
J.~Huang and G.~Huang, ``Bevdet4d: Exploit temporal cues in multi-camera 3d object detection,'' 2022. [Online]. Available: \url{https://arxiv.org/abs/2203.17054}
\BIBentrySTDinterwordspacing

\bibitem{li2022bevstereoenhancingdepthestimation}
\BIBentryALTinterwordspacing
Y.~Li, H.~Bao, Z.~Ge, J.~Yang, J.~Sun, and Z.~Li, ``Bevstereo: Enhancing depth estimation in multi-view 3d object detection with dynamic temporal stereo,'' 2022. [Online]. Available: \url{https://arxiv.org/abs/2209.10248}
\BIBentrySTDinterwordspacing

\bibitem{ho2020denoisingdiffusionprobabilisticmodels}
\BIBentryALTinterwordspacing
J.~Ho, A.~Jain, and P.~Abbeel, ``Denoising diffusion probabilistic models,'' 2020. [Online]. Available: \url{https://arxiv.org/abs/2006.11239}
\BIBentrySTDinterwordspacing

\bibitem{zhou2025detrackinmodellatentdenoising}
\BIBentryALTinterwordspacing
X.~Zhou, J.~Li, L.~Hong, K.~Jiang, P.~Guo, W.~Ge, and W.~Zhang, ``Detrack: In-model latent denoising learning for visual object tracking,'' 2025. [Online]. Available: \url{https://arxiv.org/abs/2501.02467}
\BIBentrySTDinterwordspacing

\bibitem{tian2023occ3d}
X.~Tian, T.~Jiang, L.~Yun, Y.~Wang, Y.~Wang, and H.~Zhao, ``Occ3d: A large-scale 3d occupancy prediction benchmark for autonomous driving,'' \emph{arXiv preprint arXiv:2304.14365}, 2023.

\bibitem{cao2022monoscene}
A.-Q. Cao and R.~de~Charette, ``Monoscene: Monocular 3d semantic scene completion,'' in \emph{Proceedings of the IEEE/CVF Conference on Computer Vision and Pattern Recognition}, 2022, pp. 3991--4001.

\bibitem{zhu2020deformable}
X.~Zhu, W.~Su, L.~Lu, B.~Li, X.~Wang, and J.~Dai, ``Deformable detr: Deformable transformers for end-to-end object detection,'' \emph{arXiv preprint arXiv:2010.04159}, 2020.

\bibitem{ye2021drinet++}
M.~Ye, R.~Wan, S.~Xu, T.~Cao, and Q.~Chen, ``Drinet++: Efficient voxel-as-point point cloud segmentation,'' \emph{arXiv preprint arXiv:2111.08318}, 2021.

\bibitem{tang2020searching}
H.~Tang, Z.~Liu, S.~Zhao, Y.~Lin, J.~Lin, H.~Wang, and S.~Han, ``Searching efficient 3d architectures with sparse point-voxel convolution,'' in \emph{Computer Vision--ECCV 2020: 16th European Conference, Glasgow, UK, August 23--28, 2020, Proceedings, Part XXVIII}.\hskip 1em plus 0.5em minus 0.4em\relax Springer, 2020, pp. 685--702.

\bibitem{sitzmann2019deepvoxels}
V.~Sitzmann, J.~Thies, F.~Heide, M.~Nie{\ss}ner, G.~Wetzstein, and M.~Zollhofer, ``Deepvoxels: Learning persistent 3d feature embeddings,'' in \emph{Proceedings of the IEEE/CVF Conference on Computer Vision and Pattern Recognition}, 2019, pp. 2437--2446.

\bibitem{li2022bevstereo}
Y.~Li, H.~Bao, Z.~Ge, J.~Yang, J.~Sun, and Z.~Li, ``Bevstereo: Enhancing depth estimation in multi-view 3d object detection with dynamic temporal stereo,'' \emph{arXiv preprint arXiv:2209.10248}, 2022.

\bibitem{huang2021bevdet}
J.~Huang, G.~Huang, Z.~Zhu, and D.~Du, ``Bevdet: High-performance multi-camera 3d object detection in bird-eye-view,'' \emph{arXiv preprint arXiv:2112.11790}, 2021.

\bibitem{li2022bevdepth}
Y.~Li, Z.~Ge, G.~Yu, J.~Yang, Z.~Wang, Y.~Shi, J.~Sun, and Z.~Li, ``Bevdepth: Acquisition of reliable depth for multi-view 3d object detection,'' \emph{arXiv preprint arXiv:2206.10092}, 2022.

\bibitem{FBBEV}
Z.~Li, Z.~Yu, W.~Wang, A.~Anandkumar, T.~Lu, and J.~M. Alvarez, ``Fb-bev: Bev representation from forward-backward view transformations,'' in \emph{ICCV}, 2023, pp. 6919--6928.

\bibitem{BEV-LaneDet}
R.~Wang, J.~Qin, K.~Li, Y.~Li, D.~Cao, and J.~Xu, ``Bev-lanedet: a simple and effective 3d lane detection baseline,'' in \emph{CVPR}, 2023.

\bibitem{bevpoolv2}
J.~Huang and G.~Huang, ``Bevpoolv2: A cutting-edge implementation of bevdet toward deployment,'' \emph{arXiv preprint arXiv:2211.17111}, 2022.

\bibitem{fastBEV}
B.~Huang, Y.~Li, E.~Xie, F.~Liang, L.~Wang, M.~Shen, F.~Liu, T.~Wang, P.~Luo, and J.~Shao, ``Fast-bev: Towards real-time on-vehicle bird's-eye view perception,'' \emph{arXiv preprint arXiv:2301.07870}, 2023.

\bibitem{lin2017focal}
T.-Y. Lin, P.~Goyal, R.~Girshick, K.~He, and P.~Doll{\'a}r, ``Focal loss for dense object detection,'' in \emph{Proceedings of the IEEE international conference on computer vision}, 2017, pp. 2980--2988.

\bibitem{berman2018lovasz}
M.~Berman, A.~R. Triki, and M.~B. Blaschko, ``The lov{\'a}sz-softmax loss: A tractable surrogate for the optimization of the intersection-over-union measure in neural networks,'' in \emph{Proceedings of the IEEE conference on computer vision and pattern recognition}, 2018, pp. 4413--4421.

\bibitem{caesar2020nuscenes}
H.~Caesar, V.~Bankiti, A.~H. Lang, S.~Vora, V.~E. Liong, Q.~Xu, A.~Krishnan, Y.~Pan, G.~Baldan, and O.~Beijbom, ``nuscenes: A multimodal dataset for autonomous driving,'' in \emph{Proceedings of the IEEE/CVF conference on computer vision and pattern recognition}, 2020, pp. 11\,621--11\,631.

\bibitem{wang2022internimage}
W.~Wang, J.~Dai, Z.~Chen, Z.~Huang, Z.~Li, X.~Zhu, X.~Hu, T.~Lu, L.~Lu, H.~Li, \emph{et~al.}, ``Internimage: Exploring large-scale vision foundation models with deformable convolutions,'' \emph{arXiv preprint arXiv:2211.05778}, 2022.

\bibitem{he2016deep}
K.~He, X.~Zhang, S.~Ren, and J.~Sun, ``Deep residual learning for image recognition,'' in \emph{Proceedings of the IEEE conference on computer vision and pattern recognition}, 2016, pp. 770--778.

\bibitem{zheng2015improving}
H.~Zheng, Z.~Yang, W.~Liu, J.~Liang, and Y.~Li, ``Improving deep neural networks using softplus units,'' in \emph{2015 International joint conference on neural networks (IJCNN)}.\hskip 1em plus 0.5em minus 0.4em\relax IEEE, 2015, pp. 1--4.

\bibitem{loshchilov2017decoupled}
I.~Loshchilov and F.~Hutter, ``Decoupled weight decay regularization,'' \emph{arXiv preprint arXiv:1711.05101}, 2017.

\bibitem{wang2021fcos3d}
T.~Wang, X.~Zhu, J.~Pang, and D.~Lin, ``Fcos3d: Fully convolutional one-stage monocular 3d object detection,'' in \emph{Proceedings of the IEEE/CVF International Conference on Computer Vision}, 2021, pp. 913--922.

\end{thebibliography}


\end{document}